# Convergent Propagation Algorithms via Oriented Trees

**Amir Globerson**
CSAIL
Massachusetts Institute of Technology
Cambridge, MA 02139

**Tommi Jaakkola**
CSAIL
Massachusetts Institute of Technology
Cambridge, MA 02139

## Abstract

Inference problems in graphical models are often approximated by casting them as constrained optimization problems. Message passing algorithms, such as belief propagation, have previously been suggested as methods for solving these optimization problems. However, there are few convergence guarantees for such algorithms, and the algorithms are therefore not guaranteed to solve the corresponding optimization problem. Here we present an oriented tree decomposition algorithm that is guaranteed to converge to the global optimum of the Tree-Reweighted (TRW) variational problem. Our algorithm performs local updates in the convex dual of the TRW problem – an unconstrained generalized geometric program. Primal updates, also local, correspond to oriented reparametrization operations that leave the distribution intact.

## 1 Introduction

The problem of probabilistic inference in graphical models refers to the task of calculating marginal distributions or the most likely assignment variables. Both these problems are generally NP hard, requiring approximate methods.

Many approximate inference methods, including message passing algorithms, can be viewed as trying to solve a variational formulation of the inference problem. The idea in variational approaches is to cast approximate inference as a constrained minimization of a free energy function (see [14] for a recent review). Two key questions arise in this context. The first is how to choose the free energy, and the second is how to design efficient algorithms that minimize it. When the Bethe free energy is used, it has been shown [16] that fixed-points of the belief propagation (BP) algorithm correspond to local minima of the free energy. However, BP is not generally guaranteed to converge to a fixed-point. Although there do exist algorithms that are guaranteed to converge to a local minimum of the Bethe free energy [15, 17], its global minimization is still a hard non-convex problem for which no efficient algorithms are known.

The difficulties with the Bethe free energy derive from its non-convexity and corresponding local minima problem. To avoid this difficulty, several authors have recently studied *convex* free energies [6, 7, 13]. The associated convex optimization problems can in principle be solved using generic convex optimization procedures [1] with guarantees of finding the global optimum in polynomial time. Although this presents a significant improvement over the non-convex case, the generic optimization route may be very costly in large practical problems. For example, when using a generic convex solver, every update of the variables has complexity $O(n)$, where $n$ is the number of variables. In contrast, the optimization using message passing algorithms can be reduced to local updates with $O(1)$ operations. Interestingly, even in the convex setting, the convergence of these message passing algorithms is typically not guaranteed, and damping heuristics are required to ensure convergence in practice [13]. A prominent exception is [7] where the author provides a provably convergent message passing algorithm for free energies where the entropy term is a non-negative combination of joint entropies.

Here we provide a provably convergent message passing algorithm for a specific variational setup, namely the Tree-Reweighted (TRW) optimization problem of Wainwright et al. [13]. The algorithm we propose is guaranteed to converge to the global optimum of the free energy, and does not require additional parameters such as the damping ratio. A key step in obtaining the updates is deriving the convex-dual of TRW, which we show to be an *unconstrained* instance of a generalized geometric program (GP) [3]. We derive a message passing algorithm, which we call TRW Geometric Programming (TRW-GP), that yields monotone improvement of the dual GP. We demonstrate the utility of our TRW-GP algorithm by providing an example where the TRW message passing algorithm in [13] does not converge, but TRW-GP does.

## 2 The Tree-Reweighting Formulation

We consider pairwise Markov random fields (MRF) over a set of variables $\mathbf{x} = x_1, \ldots, x_n$. Given a graph $G$ with $n$ vertices $V$ and a set of edges $E$, an MRF is a distribution over $\mathbf{x}$ defined by

$$p(\mathbf{x}; \boldsymbol{\theta}) = \frac{1}{Z(\boldsymbol{\theta})} e^{\sum_{ij \in E} \theta_{ij}(x_i, x_j) + \sum_{i \in V} \theta_i(x_i)} \quad (1)$$

where $\theta_{ij}(x_i, x_j)$ and $\theta_i(x_i)$ are parameters, $\boldsymbol{\theta}$ denotes all the parameters, and $Z(\boldsymbol{\theta})$ is the partition function.

Our focus here is on approximating singleton marginals of $p(\mathbf{x}; \boldsymbol{\theta})$, namely $p(x_i; \boldsymbol{\theta})$. This problem is closely related to that of evaluating the partition function $Z(\boldsymbol{\theta})$. We focus on the TRW variational problem which yields an upper bound on $Z(\boldsymbol{\theta})$ as well as a set of approximate marginals obtained from the minimizing solution.

We begin by briefly reviewing the TRW formalism. Consider a set of $k$ spanning trees on $G$ denoted by $T_1, \ldots, T_k$, and a distribution $\rho_i$ over these trees where $\rho_i \geq 0$ and $\sum \rho_i = 1$. To avoid overloading notation in subsequent analysis, we assume here that the trees are directed, so that the same tree structure may appear multiple times with different edge orientations. This differs from the presentation in [13] though the distinction is immaterial in the remainder of this section.

We also introduce the notion of *pseudomarginals* defined as the singleton and pairwise marginals $\mu_i(x_i), \mu_{ij}(x_i, x_j)$ associated with the edges and nodes of $G$. We use $\boldsymbol{\mu}$ to denote the set of all these marginals and $\mathcal{C}(G)$ the set of $\boldsymbol{\mu}$'s that are *pairwise consistent*

$$\sum_{x_j} \mu_{ij}(x_i, x_j) = \mu_i(x_i) \quad , \quad \sum_{x_i} \mu_{ij}(x_i, x_j) = \mu_j(x_j)$$
$$\sum_{x_i} \mu_i(x_i) = 1 \quad , \quad \mu_{ij}(x_i, x_j) \geq 0 .$$

For a given tree $T$ and $\boldsymbol{\mu} \in \mathcal{C}(G)$, define the entropy $H(\boldsymbol{\mu}; T)$ to be the entropy of an MRF on the tree $T$ with marginals given by $\boldsymbol{\mu}$. Note that only a subset of the pairwise distributions in $\boldsymbol{\mu}$ will be used for each tree, namely $\mu_{ij}(x_i, x_j)$ such that $ij$ is an edge in $T$. The tree entropy may be written in closed form as (cf. [13])

$$H(\boldsymbol{\mu}; T) = \sum_i H(X_i) - \sum_{ij \in T} I(X_i; X_j) \quad (2)$$

where $H(X_i)$ is the entropy of $\mu_i(x_i)$ and $I(X_i; X_j)$ is the mutual information calculated from $\mu_{ij}(x_i, x_j)$. Note that this expression is independent of the direction of the edges in the tree. We will make use of the directed edges in the next section.

Define the following variational *free energy* function $\mathcal{F}(\boldsymbol{\mu}; \rho, \boldsymbol{\theta})$

$$\mathcal{F}(\boldsymbol{\mu}; \rho, \boldsymbol{\theta}) = -\boldsymbol{\mu} \cdot \boldsymbol{\theta} - \sum_{i=1}^k \rho_i H(\boldsymbol{\mu}; T_i) . \quad (3)$$

In [13] it is shown that minimizing $\mathcal{F}(\boldsymbol{\mu}; \rho, \boldsymbol{\theta})$ results in an upper bound on the log-partition function

$$\log Z(\boldsymbol{\theta}) \leq - \min_{\boldsymbol{\mu} \in \mathcal{C}(G)} \mathcal{F}(\boldsymbol{\mu}; \rho, \boldsymbol{\theta}) . \quad (4)$$

The minimization also results in an optimal (minimizing) $\boldsymbol{\mu}$, which is used to approximate the marginals of $p(\mathbf{x}; \boldsymbol{\theta})$. Empirical results in [13] show that TRW usually performs as well as, and often better than the standard Bethe free energy approximations, especially in regimes where BP fails to converge.

## 3 Conditional Entropies and Directed Edge Probabilities

Our goal is to use convex duality to obtain the dual problem of Eq. (4). To achieve this, we first seek a representation of $\mathcal{F}(\boldsymbol{\mu}; \rho, \boldsymbol{\theta})$ that is a convex function of $\boldsymbol{\mu}$ for all values of $\boldsymbol{\mu}$, and not just within the consistent set $\boldsymbol{\mu} \in \mathcal{C}(G)$. For example, the entropy term in Eq. (2) is concave only for $\boldsymbol{\mu} \in \mathcal{C}(G)$ but not for a general $\boldsymbol{\mu}$. We therefore seek an alternative expression for the tree entropy.

Let $r(T)$ be the root node of $T$ (recall that the trees are directed). We write the entropy associated with the tree as

$$H(\boldsymbol{\mu}; T) = H(X_{r(T)}) + \sum_{j \to i \in T} H(X_i | X_j) \quad (5)$$

where $j \to i \in T$ implies that there is a directed edge from vertex $j$ to vertex $i$ in the directed tree $T$. The conditional entropy $H(X_i | X_j)$ is assumed to be calculated only on the basis of the *joint* marginal $\mu_{ij}(x_i, x_j)$, and does not involve $\mu_i(x_i)$. The entropy $H(X_{r(T)})$ is calculated via the singleton marginal $\mu_{r(T)}(x_{r(T)})$. The expressions in Eq. (5) and Eq. (2) will agree whenever $\boldsymbol{\mu} \in \mathcal{C}(G)$. However, they will yield different results when $\boldsymbol{\mu} \notin \mathcal{C}(G)$.

The advantage of Eq. (5) is that $H(\boldsymbol{\mu}; T)$ is now a concave function of the set of marginals $\boldsymbol{\mu}$. The concavity follows immediately from the concavity of $H(X_i)$ as a function of $\mu_i(x_i)$ and the concavity of the conditional entropy $H(X_i | X_j)$ as a function of $\mu_{ij}(x_i, x_j)$ [6].

The function $\mathcal{F}(\boldsymbol{\mu}; \rho, \boldsymbol{\theta})$ involves a summation over a potentially large number of tree entropies. To express this compactly while maintaining directionality, we define $\rho_{i|j}$ as the probability that the directed edge $j \to i$ is present in a tree drawn according to the distribution $\rho$ over trees. Similarly, we define $\rho_{oi}$ as the probability that node $i$ appears as a root. We note that it is possible to find such edge probabilities for distributions (e.g. uniform) over the set of *all* spanning trees by employing a variant of the matrix tree theorem for directed trees (see [12] p. 141 and [11]).

The function $\mathcal{F}(\boldsymbol{\mu}; \rho, \boldsymbol{\theta})$ can now be written as

$$-\boldsymbol{\mu} \cdot \boldsymbol{\theta} - \sum_{i \in V} \rho_{oi} H(X_i) - \sum_{ij \in \bar{E}} \rho_{i|j} H(X_i | X_j) \quad (6)$$

where the edge set $\bar{E}$ contains edges in both directions. In other words, if $ij \in \bar{E}$ then $ji$ is also in $\bar{E}$. The new function $\mathcal{F}(\boldsymbol{\mu}; \rho, \boldsymbol{\theta})$ is convex in $\boldsymbol{\mu}$ without assuming consistency of the marginals.

## 4  The TRW Convex Dual

The TRW primal problem is given by

$$\min_{\boldsymbol{\mu} \in \mathcal{C}(G)} \mathcal{F}(\boldsymbol{\mu}; \rho, \boldsymbol{\theta}) \ . \quad (7)$$

Since the function $\mathcal{F}(\boldsymbol{\mu}; \rho, \boldsymbol{\theta})$ is now convex for *all* $\boldsymbol{\mu}$ and the set of constraints is linear, this optimization problem is convex and thus has an equivalent convex dual [1].[1] However, it is not immediately clear how to derive this dual in closed form. The main difficulty is that two terms in the objective $\mathcal{F}(\boldsymbol{\mu}; \rho, \boldsymbol{\theta})$ depend on $\mu_{ij}(x_i, x_j)$, namely $H(X_i | X_j)$ and $H(X_j | X_i)$. To get around this problem we introduce additional variables to the primal problem. Specifically, we replace $\mu_{ij}(x_i, x_j)$ by two *copies* which we denote by $\mu_{i|j}(x_i, x_j)$ and $\mu_{j|i}(x_i, x_j)$, and require that these two copies are identical. The entropy $H(X_i | X_j)$ is then evaluated via the variables $\mu_{i|j}(x_i, x_j)$. We shall also find it convenient to replace the consistency constraints in $\mathcal{C}(G)$ by the following equivalent *directed* consistency constraints

$$\mu_{i|j}(x_i, x_j) = \mu_{j|i}(x_i, x_j)$$
$$\sum_{x_j} \mu_{j|i}(x_i, x_j) = \mu_i(x_i) \quad , \quad \sum_{x_i} \mu_{i|j}(x_i, x_j) = \mu_j(x_j)$$
$$\sum_{x_i} \mu_i(x_i) = 1 \quad , \quad \mu_{i|j}(x_i, x_j) \geq 0 \ , \ \mu_{j|i}(x_i, x_j) \geq 0 \ .$$

For simplicity we will continue to denote the new extended variable set by $\boldsymbol{\mu}$ (as we will be using it from

now on) and refer to the consistency constraints by $\vec{\mathcal{C}}(G)$. The TRW primal problem is then

$$\underline{PTRW}: \quad \min_{\boldsymbol{\mu} \in \vec{\mathcal{C}}(G)} \mathcal{F}(\boldsymbol{\mu}; \rho, \boldsymbol{\theta}) \ . \quad (8)$$

The convex dual of $PTRW$ is derived in App. A. and is in fact a convex unconstrained minimization problem. In what follows we describe this dual. The dual variables will be denoted by $\beta_{ij}(x_i, x_j)$ for $ij \in E$, and are not constrained.[2] The dual objective is given by

$$\mathcal{F}_D(\boldsymbol{\beta}; \rho, \boldsymbol{\theta}) = \sum_i \rho_{oi} \log \sum_{x_i} e^{\rho_{oi}^{-1}(\theta_i(x_i) - \sum_{k \in N(i)} \lambda_{k|i}(x_i; \boldsymbol{\beta}))}$$

where $\lambda_{j|i}(x_i; \boldsymbol{\beta})$ is a function of the $\boldsymbol{\beta}$ variables:

$$\lambda_{j|i}(x_i; \boldsymbol{\beta}) = -\rho_{j|i} \log \sum_{x_j} e^{\rho_{j|i}^{-1}(\theta_{ij}(x_i, x_j) + \delta_{j|i} \beta_{ij}(x_i, x_j))}$$

and $\delta_{j|i}$ is defined as

$$\delta_{j|i} = \begin{cases} 1 & ji \in E \\ -1 & ij \in E \end{cases} .$$

The dual TRW optimization problem is then

$$\underline{DTRW}: \quad \min_{\boldsymbol{\beta}} \mathcal{F}_D(\boldsymbol{\beta}; \rho, \boldsymbol{\theta}) \quad . \quad (9)$$

We re-emphasize the fact the DTRW is an unconstrained minimization of a function of $\boldsymbol{\beta}$. The variables $\lambda_{j|i}(x_i; \boldsymbol{\beta})$ are introduced merely for the purpose of notational convenience. The mapping between dual and primal variables can be shown to be

$$\mu_i(x_i) \propto e^{\rho_{oi}^{-1}(\theta_i(x_i) - \sum_{k \in N(i)} \lambda_{k|i}(x_i; \boldsymbol{\beta}))}$$
$$\mu_{j|i}(x_j | x_i) \propto e^{\rho_{j|i}^{-1}(\theta_{ij}(x_i, x_j) + \delta_{j|i} \beta_{ij}(x_i, x_j))} \quad . \quad (10)$$

This relation maps the optimal $\boldsymbol{\beta}$ to the optimal $\boldsymbol{\mu}$, but we shall also use it for non-optimal values.

The dual objective $\mathcal{F}_D(\boldsymbol{\beta}; \rho, \boldsymbol{\theta})$ is a convex function (see App. B) and therefore has no local minima.

## 5  Dual Gradient and Optimum

The DTRW problem presented above is unconstrained and can thus be solved using a variety of gradient based algorithms, such as conjugate gradient or BFGS [10]. The gradient of $\mathcal{F}_D(\boldsymbol{\beta}; \rho, \boldsymbol{\theta})$ w.r.t. $\boldsymbol{\beta}$ is

$$\frac{\partial \mathcal{F}_D(\boldsymbol{\beta}; \rho, \boldsymbol{\theta})}{\partial \beta_{ij}(x_i, x_j)} = \mu_{i|j}(x_i | x_j) \mu_j(x_j) - \mu_{j|i}(x_j | x_i) \mu_i(x_i)$$

where the distributions are given by the dual to primal mapping in Eq. (10). The gradient is thus a measure of the discrepancy between two ways of calculating the joint pairwise marginal, based on the two different orientations of the edge $ij$.

---

[1] Strict duality follows from Slater's conditions, which are satisfied in this case.

[2] Note that $\beta$ variables are not *directed*, i.e., there is one variable $\beta_{ij}$ per edge.

To characterize the optimum of DTRW we set the gradient to zero, yielding the following simple dual optimality criterion

$$\mu_{i|j}(x_i|x_j)\mu_j(x_j) = \mu_{j|i}(x_j|x_i)\mu_i(x_i) \ . \quad (11)$$

Thus at the optimum the two alternative ways of estimating $\mu_{ij}(x_i, x_j)$ will yield the same result.

Calculating the gradient w.r.t a given $\beta_{ij}(x_i, x_j)$ has complexity $O(1)$, and relies only on $\beta_{ij}(x_i, x_j)$ for edges containing $i$ or $j$. Thus the gradient can be calculated locally, and gradient descent algorithms can be implemented efficiently. One drawback of gradient based algorithms is their reliance on line-search modules for finding a step size that decreases the objective. In the next section we consider updates that are parameter-free.

## 6 Local Marginal Updates

The gradient updates described in the previous section use the *difference* between two joint distributions. We will now focus on updates relying on the *ratio* between these distributions. Consider

$$\beta_{ij}^{t+1}(x_i, x_j) = \beta_{ij}^t(x_i, x_j) + \epsilon \log \frac{\mu_{j|i}^t(x_j|x_i)\mu_i^t(x_i)}{\mu_{i|j}^t(x_i|x_j)\mu_j^t(x_j)} \quad (12)$$

where $\mu_{j|i}^t(x_j|x_i)$ and $\mu_i^t(x_i)$ are functions of $\beta$ as in Eq. (10) and $\epsilon$ is a step size whose value will be discussed in the next section. As a ratio of two expected values, the update is reminiscent of Generalized Iterative Scaling [5]. We shall assume for simplicity that only one edge is updated at each time step $t$.

The update in Eq. (12) is performed on the $\beta$ variables. An equivalent, and somewhat simpler update may be derived in terms of the variables $\mu_{i|j}^t(x_i|x_j)$ and $\mu_j^t(x_j)$. The resulting updates and algorithm are described in Figure 1. We call the resulting algorithm TRW-GP (TRW Geometric Programming).

### 6.1 Convergence Proof

To analyze the convergence of the update in the previous section, we need to consider the resulting change in the objective $\mathcal{F}_D(\beta; \rho, \theta)$, namely $\mathcal{F}_D(\beta^t; \rho, \theta) - \mathcal{F}_D(\beta^{t+1}; \rho, \theta)$. It can be shown (see App. D) that this difference only depends on the $\mu$ variables in the TRW-GP algorithm, and thus we denote it by $\Delta_D(\mu^t)$. Since $\mathcal{F}_D(\beta^t; \rho, \theta)$ should be minimized, this difference needs to be non-negative. This is indeed guaranteed by the following lemma (see App. D):

**Lemma 6.1:** *For $0 < \epsilon < \min(\rho_{oi}, \rho_{oj}, \rho_{i|j}, \rho_{j|i})$ the dual objective is decreased at every iteration so that $\Delta_D(\mu^t) \geq 0$ for all $t$. Furthermore, $\Delta_D(\mu^t) = 0$ holds if and only if the optimum condition of Eq. (11) is satisfied.*

Any choice of $\epsilon$ that is smaller than $\min(\rho_{oi}, \rho_{oj}, \rho_{i|j}, \rho_{j|i})$ will result in monotone improvement of the objective. In the current implementation we use $\epsilon = \frac{1}{2}\min(\rho_{oi}, \rho_{oj}, \rho_{i|j}, \rho_{j|i})$. This value turns out to minimize a first order approximation of the improvement in the objective, and was found to work well in practice. The convergence to the global optimum now follows from Lemma 6.1.

**Lemma 6.2:** *The updates in Eq. (12) with $\epsilon$ as in Lemma 6.1 converge to the joint optimum of PTRW and DTRW.*

**Proof:** Denote the mapping from $\mu^t$ to $\mu^{t+1}$ by $\mathcal{R}(\mu^t) = \mu^{t+1}$. The mapping is clearly continuous. By Lemma 6.1 the sequence $\mathcal{F}_D(\beta^t; \rho, \theta)$ is monotonically decreasing. It is also bounded since $\mathcal{F}_D(\beta; \rho, \theta)$ is bounded and thus the difference series $\Delta_D(\mu^t)$ converges to zero. Taking $t$ to infinity then implies that $\mu^t$ has a convergent subsequence that converges to some $\mu^*$. This $\mu^*$ will then satisfy $\mathcal{F}_D(\mu^*; \rho, \theta) = \mathcal{F}_D(\mathcal{R}(\mu^*); \rho, \theta)$. We know from the Lemma 6.1 that such a point necessarily satisfies the zero gradient condition in Eq. (11), and thus $\mu^*$ (or more precisely, the corresponding $\beta$) minimizes the dual objective.[3] $\quad\square$

## 7 Tree Re-parametrization View

The TRW problem can be interpreted in terms of iterating through different re-parametrizations of the distribution $p(\mathbf{x}; \theta)$ [13]. Here we present a related view of our algorithm.

We wish to show that the marginal variables obtained by the algorithm can always be used to obtain the original distribution via

$$p(\mathbf{x}; \theta) = c_t \prod_i \mu_i^t(x_i)^{\rho_{oi}} \prod_{ij \in \bar{E}} \mu_{j|i}^t(x_j|x_i)^{\rho_{j|i}} \ . \quad (13)$$

For $t = 0$ this is clearly true. We proceed by induction. Assume that at iteration $t$ we have a reparametrization with constant $c_t$. Substituting the update rule in Figure 1 and using simple algebra shows that we again have a reparametrization, only with

$$c_{t+1} = c_t e^{\mathcal{F}_D(\beta^{t+1}; \rho, \theta) - \mathcal{F}_D(\beta^t; \rho, \theta)} = c_t e^{-\Delta_D(\mu^t)} \ .$$

---

[3]To carefully account for the possibility that some of the converging marginals would involve zero probabilities, the updates in the primal form, along with the objective, can be written in a form without any ratios.

---

**Inputs:** A graph $G = (E, V)$, parameter vector $\boldsymbol{\theta}$ on $G$, root probabilities $\rho_{\circ i}$ and directed edge probabilities $\rho_{i|j}$ for $(ij), (ji) \in E$.

**Initialization:** Set $\mu_i^0(x_i) \propto e^{\rho_{\circ i}^{-1} \theta_i(x_i)}$ and $\mu_{i|j}^0(x_i|x_j) \propto e^{\rho_{i|j}^{-1} \theta_{ij}(x_i, x_j)}$

**Algorithm:** Iterate until small enough change in marginals:

- Set $\epsilon = \frac{1}{2} \min(\rho_{\circ i}, \rho_{\circ j}, \rho_{i|j}, \rho_{j|i})$, and update

$$\mu_i^{t+1}(x_i) \propto \mu_i^t(x_i) \left( \sum_{x_j} \mu_{j|i}^t(x_j|x_i) \left( \frac{\mu_{i|j}^t(x_i|x_j)\mu_j^t(x_j)}{\mu_{j|i}^t(x_j|x_i)\mu_i^t(x_i)} \right)^{\epsilon \rho_{j|i}^{-1}} \right)^{\rho_{j|i}\rho_{\circ i}^{-1}}$$

$$\mu_{i|j}^{t+1}(x_i|x_j) \propto \mu_{i|j}^t(x_i|x_j)^{1 - \epsilon \rho_{i|j}^{-1}} \left( \frac{\mu_{j|i}^t(x_j|x_i)\mu_i^t(x_i)}{\mu_j^t(x_j)} \right)^{\epsilon \rho_{i|j}^{-1}}$$

**Output:** Final values of marginals.

---

Figure 1: The TRW-GP algorithm expressed in terms of conditional and singleton marginals.

In other words the multiplicative constant turns out to be related to the improvement in the dual function. This creates an interesting link between reparametrization and minimization, and may be used to study message passing algorithms where a dual is more difficult to characterize.

## 8 Relation to Previous Work

Heskes [7] recently presented a detailed study of convex free energies. When the entropy term is a positive combination of joint and singleton entropies (and is therefore concave), he provides a local update algorithm that is monotone in the convex dual, and converges to the global optimum. He then discusses the application of the same algorithm to the case where the singleton entropies all have negative weight, and the overall entropy is convex over the set of constraints.[4] In this case, the dual is generally not given in closed form and it is not known if the algorithm decreases it at every step. However, Heskes argues that with sufficient damping the algorithm can be shown to converge, although the exact form of damping is not given.

Since the TRW entropy can be shown to decompose into positively weighted pairwise entropies and negatively weighted singleton entropies, it satisfies the above conditions in Heskes' work. Our analysis provides several advantages over the algorithm in [7]. First, we derive a closed form solution of dual. Second, the dual is unconstrained, and thus allows unconstrained minimization methods to be applied. Third, unlike most belief propagation variants, our algorithm is shown to provide a monotone improvement of an objective function[5], and thus diverges from the standard fixed point analysis used in message passing algorithms.

Finally, another algorithm which is guaranteed to converge to a global minimum of convexified free energies is the double loop CCCP algorithm of Yuille [17]. The main disadvantage of CCCP is that each iteration requires solving an optimization problem. This usually results in slower convergence, and furthermore it is not clear what precision is required for the inner loop optimization, and how this affects convergence guarantees. The algorithm we present here is essentially a *single loop* method, and is thus easier to analyze.

## 9 Empirical Demonstration

The original TRW message passing (TRW-MP) algorithm presented in [13] is not generally guaranteed to converge. However, we observed empirically that when damping of $\alpha = 0.5$ is applied to the log-messages, convergence is always achieved.[6] To compare TRW-MP to TRW-GP, we use the pseudomarginals generated by TRW-MP[7] as marginals in the primal objective $\mathcal{F}(\boldsymbol{\mu}; \rho, \boldsymbol{\theta})$ in Eq. (3). This value is not expected to be an upper or lower bound on the optimum of $\mathcal{F}(\boldsymbol{\mu}; \rho, \boldsymbol{\theta})$, since the TRW-MP pseudomarginals are not guaranteed to be pairwise consistent, except at the optimum. However, since the TRW-MP pseudo-marginals converge to the optimal primal marginals, the value $\mathcal{F}(\boldsymbol{\mu}; \rho, \boldsymbol{\theta})$ will converge to the primal optimum. The progress of TRW-GP may be monitored by evaluating $\mathcal{F}_D(\boldsymbol{\beta}; \rho, \boldsymbol{\theta})$ at every iteration. This value is guaranteed to decrease and converge to the optimum of $\mathcal{F}_D(\boldsymbol{\beta}; \rho, \boldsymbol{\theta})$ which is identical to the optimum of $\mathcal{F}(\boldsymbol{\mu}; \rho, \boldsymbol{\theta})$. We can thus observe the rate at which the different algorithms converge to their joint optimum.

---

[4] The discussion in [7] is in terms of general regions, not just pairs. We present his argument for the simpler pairwise case.

[5] As mentioned above, Heskes presents such an algorithm for positively weighted singleton *and* pairwise entropies. It is however not clear that such entropies are useful in practice

[6] This observation is in line with Heskes' argument that sufficiently damped messages will converge for the case of the TRW free energy.

[7] See Equations (58) and (59) in [13].

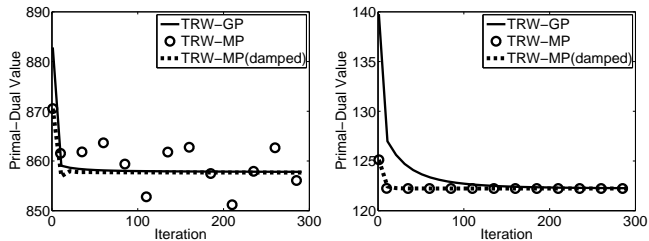

Figure 2: Illustration of the dual message passing algorithm for a $10 \times 10$ Ising model. The TRW-GP curve shows the dual objective value $\mathcal{F}_D(\boldsymbol{\beta}; \rho, \boldsymbol{\theta})$ obtained by the TRW-GP algorithm. The TRW-MP curves show the primal objective values $\mathcal{F}(\boldsymbol{\mu}; \rho, \boldsymbol{\theta})$ obtained by TRW message passing algorithms. The damped TRW-MP used a damping of 0.5 in the log domain. The MRF parameters were set as follows: $\alpha_F = 1, \alpha_I = 9$ for the left figure, and $\alpha_F = 1, \alpha_I = 1$ for the right figure.

To study the convergence rate of the two algorithms, we used an Ising model on a $10 \times 10$ grid with interaction parameters $\theta_{ij}$ drawn uniformly from $[-\alpha_I, \alpha_I]$ and field parameters $\theta_i$ drawn uniformly from $[-\alpha_F, \alpha_F]$. The MRF is given by $p(\mathbf{x}; \theta) \propto e^{\sum_{ij \in E} \theta_{ij} x_i x_j + \sum_{i \in V} \theta_i x_i}$ where $x_i \in \{+1, -1\}$. We used a uniform distribution over directed spanning trees calculated as in [11].

Figure 2 (left) shows an example run where the undamped TRW-MP algorithm does not converge, but the TRW-GP and the damped TRW-MP do converge, and do so roughly at the same rate. Figure 2 (right) shows an example where both TRW-MP algorithms converge and do so at a faster rate than TRW-GP. We experimented with various values of $\alpha_F$ and $\alpha_I$ and have observed that at lower interaction levels (e.g., $\alpha_I \leq 4$ for $\alpha_F = 1$) the TRW-MP algorithms outperform TRW-GP, whereas for higher interaction levels the undamped TRW-MP does not converge, but the damped version converges at roughly the same rate as TRW-GP. We also experimented with conjugate gradient minimization of $\mathcal{F}_D(\boldsymbol{\beta}; \rho, \boldsymbol{\theta})$, but these did not yield better rates than TRW-GP.

## 10 Conclusions

We presented a novel message passing algorithm whose updates yield a monotone improvement on the dual of the TRW free energy minimization problem. In order to obtain a closed form dual we used two tricks. The first was to decouple different entropies that depend on the same marginals by introducing multiple copies of these marginals. The second was to use uni-directional consistency constraints, so that every copy of a joint marginal appears in a single consistency constraint. Although we presented the method in the context of tree decompositions, the algorithm itself still applies as long as $\rho_{i|j}$ and $\rho_{\circ i}$ are non-negative (although the upper bound on the log partition function may not be guaranteed in this case).

The TRW-GP algorithm resolves the convergence problems with the undamped TRW-MP algorithm. However, we observed empirically that the damped TRW-MP algorithm always converges, and typically at a better rate than TRW-GP. Thus, the main contribution of the current paper is in introducing a dual framework for message passing algorithms, which could be used to analyze existing algorithms, and possibly develop faster variants in the future.

Free energies may be defined using marginals of more than two variables [13, 16]. In a recent paper [6] we study the relation between such free energies and GP. It will be worthwhile to study generalizations of TRW-GP to this case. Another interesting extension is to the MAP problem, where the corresponding variational problem is a linear program. Global convergence results for MAP message passing algorithms such as max-product are also hard to obtain in the general case. It turns out that an approach similar to the one presented here may be used to obtained convergent algorithms to solve the MAP linear program. These algorithms will be presented elsewhere.

## A    Deriving the TRW Dual

Our goal is to show that the problems in Eq. (8) and Eq. (9) are convex duals of each other.

First, we claim that the convex dual of the PTRW problem in Eq. (8) is given by

> *DTRWC*
>
> $\min \quad \sum_i \rho_{oi} \log \sum_{x_i} e^{\rho_{oi}^{-1}\left(\theta_i(x_i) - \sum_{k \in N(i)} \lambda_{k|i}(x_i)\right)}$
>
> $s.t. \quad \sum_{x_j} e^{\rho_{j|i}^{-1}(\theta_{ij}(x_i,x_j) + \delta_{j|i}\beta_{ij}(x_i,x_j) + \lambda_{j|i}(x_i))} \leq 1 \ .$

The variables in the above problem are $\lambda_{j|i}(x_i)$, $\lambda_{i|j}(x_j)$ and $\beta_{ij}(x_i, x_j)$ for every edge $ij \in E$.

The duality between PTRW and DTRWC results from the duality between conditional entropy maximization and geometric programs, and appears in several works in slightly different forms [3, 8]. A derivation of the duality result can be found in [2] (page 256) and [4]. It is important to note that the dual can be found in this case because the objective is a sum of conditional entropies (and singleton entropies) as in Eq. (6). It is not clear how to derive a dual if tree entropies are expressed via mutual information as in Eq. (2).

Due to complementary slackness conditions, the inequality in the constraints of DTRWC will hold with equality at the optimum iff the optimal primal variables satisfy $\mu_i(x_i) > 0$. In App. C we show that for the current objective, this will always happen, i.e., $\mu_i(x_i) > 0$ for all $i$ and $x_i$. We thus conclude that *all* the inequality constraints in DTRWC are always satisfied as equalities at the optimum. We therefore lose nothing by replacing them with equality constraints

$$\sum_{x_j} e^{\rho_{j|i}^{-1}(\theta_{ij}(x_i,x_j) + \delta_{j|i}\beta_{ij}(x_i,x_j) + \lambda_{j|i}(x_i))} = 1 \ . \quad (14)$$

Since each variable $\lambda_{j|i}(x_i)$ appears in only one constraint, we can eliminate it by expressing it as a function of the $\boldsymbol{\beta}$ variables

$$\lambda_{j|i}(x_i; \boldsymbol{\beta}) = -\rho_{j|i} \log \sum_{x_j} e^{\rho_{j|i}^{-1}(\theta_{ij}(x_i,x_j) + \delta_{j|i}\beta_{ij}(x_i,x_j))} \ .$$

Since the $\lambda_{j|i}(x_i)$ variables have been eliminated and the equality constraints are satisfied, optimization is now only over the $\boldsymbol{\beta}$ variables, yielding the DTRW problem in Eq. (9)

$$\min_{\boldsymbol{\beta}} \quad \sum_i \rho_{oi} \log \sum_{x_i} e^{\rho_{oi}^{-1}\left(\theta_i(x_i) - \sum_{k \in N(i)} \lambda_{k|i}(x_i; \boldsymbol{\beta})\right)} \ .$$

## B    Convexity of the Dual

Here we argue that the function $\mathcal{F}_D(\boldsymbol{\beta}; \rho, \boldsymbol{\theta})$ is a convex function of $\boldsymbol{\beta}$. We first define the class of *posynomial* functions as functions of the form [9]

$$f(x_1, \ldots, x_n) = \sum_{k=1}^K c_k x_1^{a_{1k}} x_2^{a_{2k}}, \ldots, x_n^{a_{1n}} \quad (15)$$

where $c_k > 0$. A function $f(x_1, \ldots, x_n)$ is said to be a *generalized posynomial* if it is either a posynomial or it can be formed from generalized posynomials using the operations of addition, multiplication, positive power, maximum and composition. A key property of generalized posynomials is that they can be turned into a convex function by a simple change of variables. Specifically, if $f(x)$ is a generalized posynomial, then $F(y) = \log f(e^y)$ is a convex function of $y$ [9].

It is easy to see that $f_{j|i}(x_i; e^{\boldsymbol{\beta}}) = e^{-\lambda_{j|i}(x_i; \boldsymbol{\beta})}$ is a generalized posynomial in $e^{\boldsymbol{\beta}}$ (since it is a positive power of a posynomial). The function

$$
\begin{aligned}
g_i(e^{\boldsymbol{\beta}}) &= \sum_{x_i} e^{\rho_{oi}^{-1}\left(\theta_i(x_i) - \sum_{k \in N(i)} \lambda_{k|i}(x_i; \boldsymbol{\beta})\right)} \\
&= \sum_{x_i} e^{\rho_{oi}^{-1}\theta_i(x_i)} \left(\prod_{k \in N(i)} f_{k|i}(x_i; e^{\boldsymbol{\beta}})\right)^{\rho_{oi}^{-1}}
\end{aligned}
$$

is then also a generalized posynomial. Therefore $\log g_i(e^{\boldsymbol{\beta}})$ is a convex function of $\boldsymbol{\beta}$. Since $\mathcal{F}_D(\boldsymbol{\beta}; \rho, \boldsymbol{\theta}) = \sum_i \rho_{oi} \log g_i(e^{\boldsymbol{\beta}})$, it follows that $\mathcal{F}_D(\boldsymbol{\beta}; \rho, \boldsymbol{\theta})$ is a convex function of $\boldsymbol{\beta}$.

## C  Strict Positivity of TRW Marginals

Here we want to show that the solution of the primal problem must satisfy $\mu_i(x_i) > 0$. To do so, we employ an alternative formulation of TRW [13]. Assign a parameter vector $\boldsymbol{\theta}_T$ to every tree in the set of trees, and denote by $Z(\boldsymbol{\theta}_T)$ the partition function of an MRF on tree $T$ with parameters $Z(\boldsymbol{\theta}_T)$. TRW can then be cast as

$$
\begin{aligned}
\min \quad & \sum_T \rho_T \log Z(\boldsymbol{\theta}_T) \\
s.t. \quad & \sum_T \rho_T \boldsymbol{\theta}_T = \boldsymbol{\theta} \ .
\end{aligned} \tag{16}
$$

At the optimum, all tree distributions $p(\mathbf{x}; \boldsymbol{\theta}_T)$ can be shown to have the same singleton marginals $\mu_i(x_i)$, and these correspond to the marginals that solve PTRW. The optimization above can be rewritten as

$$
\begin{aligned}
\min \quad & \sum_T \rho_T D_{KL}[p(\mathbf{x}; \boldsymbol{\theta}) | p(\mathbf{x}; \boldsymbol{\theta}_T)] \\
s.t. \quad & \sum_T \rho_T \boldsymbol{\theta}_T = \boldsymbol{\theta} \ .
\end{aligned} \tag{17}
$$

The objective in Eq. (16) can be obtained from that in Eq. (17) by expanding the $D_{KL}$ and using the fact that the constraints hold. The two objectives then differ by a constant $\log Z(\boldsymbol{\theta})$

Assume that at the optimum of PTRW, there exist $i$ and $x_i$ such that $\mu_i(x_i) = 0$. Then the above argument implies that all trees distributions will have this zero marginal. However, in that case there will be an assignment $\mathbf{x}$ such that $p(\mathbf{x}; \boldsymbol{\theta}_T) = 0$. On the other hand, for any finite $\boldsymbol{\theta}$ the true distribution $p(\mathbf{x}; \boldsymbol{\theta})$ will be strictly greater than zero. The $D_{KL}$ will then be infinite, implying the parameters are not optimal and resulting in a contradiction.[8]

## D  Monotonicity of Updates

Assume we perform an update on the $\boldsymbol{\mu}$ variables corresponding to an edge $ij \in E$ (i.e., update

$\mu_{i|j}(x_i, x_j), \mu_{j|i}(x_i, x_j), \mu_i(x_i)$ and $\mu_j(x_j)$) and that $\epsilon < \min(\rho_{oi}, \rho_{oi}, \rho_{i|j}, \rho_{j|i})$. The resulting difference in objective value can be written as $\Delta_D(\boldsymbol{\mu}^t) = f_i + f_j$ where

$$
f_i = -\rho_{oi} \log \sum_{x_i} \mu_i^t(x_i) e^{\rho_{oi}^{-1}(\lambda_{j|i}^t(x_i) - \lambda_{j|i}^{t+1}(x_i))} \ . \tag{18}
$$

The $\lambda_{j|i}^t(x_i) - \lambda_{j|i}^{t+1}(x_i)$ difference can be written in terms of $\boldsymbol{\mu}^t$ as

$$
\begin{aligned}
&\rho_{j|i} \log \sum_{x_j} \mu^t(x_j|x_i)^{1 - \epsilon \rho_{j|i}^{-1}} \left(\frac{\mu^t(x_i|x_j)\mu^t(x_j)}{\mu^t(x_i)}\right)^{\epsilon \rho_{j|i}^{-1}} = \\
&\rho_{j|i} \log \sum_{x_j} e^{(1 - \epsilon \rho_{j|i}^{-1})\log \mu^t(x_j|x_i) + \epsilon \rho_{j|i}^{-1} \log \frac{\mu^t(x_i|x_j)\mu^t(x_j)}{\mu^t(x_i)}}
\end{aligned}
$$

Since $0 < \epsilon < \rho_{j|i}$ we can use the convexity of the $\log \sum \exp$ function and the fact that $\sum_{x_j} \mu_{j|i}^t(x_j|x_i) = 1$ to obtain

$$
\lambda_{j|i}^t(x_i) - \lambda_{j|i}^{t+1}(x_i) \ \leq \ \epsilon \log \sum_{x_j} \frac{\mu_{i|j}^t(x_i|x_j)\mu_j^t(x_j)}{\mu_i^t(x_i)} \ .
$$

Note that since $\log \sum \exp$ is strictly convex, equality here is achieved if and only if $\mu^t(x_j|x_i) = \frac{\mu^t(x_i|x_j)\mu^t(x_j)}{\mu^t(x_i)}$, which implies the optimum condition in Eq. (11) is satisfied.

Substituting this in the expression for $f_i$ in Eq. (18) and rearranging we have

$$
f_i \geq -\rho_{oi} \log \sum_{x_i} (\mu_i^t(x_i)^{1 - \frac{\epsilon}{\rho_{oi}}} \left(\sum_{x_j} \mu_{i|j}^t(x_i|x_j)\mu_j^t(x_j)\right)^{\frac{\epsilon}{\rho_{oi}}}
$$

The above expression is of the form $\log \sum_{x_i} p(x_i)^{1 - \eta} q(x_i)^\eta$ where $p(x_i), q(x_i)$ are distributions over $x_i$. Define a distribution $r(x_i) \propto p(x_i)^{1 - \eta} q(x_i)^\eta$. Simple algebra then yields

$$
-\log \sum_{x_i} p(x_i)^{1 - \eta} q(x_i)^\eta = (1 - \eta) D_{KL}[r|p] + \eta D_{KL}[r|q]
$$

where $D_{KL}$ is the KL divergence, and is non-negative. Here $\eta = \frac{\epsilon}{\rho_{oi}}$ and thus $0 < \eta < 1$ and the above weighted sum of the two $D_{KL}$ divergences is always non-negative.

It follows that $f_i \geq 0$ with equality if and only if the condition in Eq. (11) is satisfied . A similar argument shows that $f_j \geq 0$ if with equality iff Eq. (11) is satisfied. The result for the non-negativity of $\Delta_D(\boldsymbol{\mu}^t)$ then follows immediately.

---

[8] We note that the same argument can be applied to show that the optimal pairwise marginals $\mu_{ij}(x_i, x_j)$ are never zero.